# Towards Dropout Training for Convolutional Neural Networks

**Haibing Wu**

Department of Electronic Engineering

Fudan University

Shanghai 200433, China

`haibingwu13@fudan.edu.cn`

**Xiaodong Gu**

Department of Electronic Engineering

Fudan University

Shanghai 200433, China

`xdgu@fudan.edu.cn`

**Abstract**

Recently, dropout has seen increasing use in deep learning. For deep convolutional neural networks, dropout is known to work well in fully-connected layers. However, its effect in convolutional and pooling layers is still not clear. This paper demonstrates that *max-pooling dropout* is equivalent to randomly picking activation based on a multinomial distribution at training time. In light of this insight, we advocate employing our proposed *probabilistic weighted pooling,* instead of commonly used max-pooling, to act as model averaging at test time. Empirical evidence validates the superiority of probabilistic weighted pooling. We also empirically show that the effect of convolutional dropout is not trivial, despite the dramatically reduced possibility of over-fitting due to the convolutional architecture. Elaborately designing dropout training simultaneously in max-pooling and fully-connected layers, we achieve state-of-the-art performance on MNIST, and very competitive results on CIFAR-10 and CIFAR-100, relative to other approaches without data augmentation. Finally, we compare max-pooling dropout and stochastic pooling, both of which introduce stochasticity based on multinomial distributions at pooling stage.

## 1 Introduction

Deep convolutional neural networks (CNNs) have recently told many success stories in visual recognition tasks and are now record holders on many challenging datasets. A standard CNN consists of alternating convolutional and pooling layers, with fully-connected layers on top. Compared to regular feed-forward networks with similarly-sized layers, CNNs have much fewer connections and parameters due to the local-connectivity and shared-filter architecture in convolutional layers, so they are far less prone to over-fitting. Another nice property of CNNs is that pooling operation provides a form of translation invariance and thus benefits generalization. Despite these attractive qualities and despite the fact that CNNs are much easier to train than other regular, deep, feed-forward neural networks, big CNNs with millions or billions of parameters still easily overfit relatively small training data.

Dropout (Hinton et al., 2012) is a recently proposed regularizer to fight against over-fitting. It is a regularization method that stochastically sets to zero the activations of hidden units for each training case at training time. This breaks up co-adaptions of feature detectors since the dropped-out units cannot influence other retained units. Another way to interpret dropout is that it yields a very efficient form of model averaging where the number of trained models is exponential in that of units, and these models share the same parameters. Dropout has also inspired other stochastic model averaging methods such as stochastic pooling (Zeiler & Fergus, 2013) and DropConnect (Wan et al., 2013).

Although dropout is known to work well in fully-connected layers of convolutional neural nets (Hinton et al., 2012; Wan et al., 2013; Krizhevsky, Sutskever, & Hinton, 2012), its effect in convolutional and pooling layers is, however, not well studied. This paper shows that using *max-pooling dropout* at training time is equivalent to sampling activation based on a multinomial distribution, and the distribution has a tunable parameter *p* (the retaining probability). In light of this, *probabilistic weighted pooling* is proposed and employed at test time to efficiently average all possibly max-pooling dropout trained networks. Our empirical evidence confirms the superiority of probabilistic weighted pooling over max-pooling. Like *fully-connected dropout*, the number of possible max-pooling dropout models also grows exponentially with the increase of the number of hidden units that are fed into pooling layers, but decreases with the increase of pooling region's size. We also empirically show that the effect of *convolutional dropout* is not trivial, despite the dramatically reduced possibility of over-fitting due to the convolutional architecture. Carefully designing dropout training simultaneously in max-pooling and fully-connected layers, we report state-of-the-art results on MNIST, and very competitive results on CIFAR-10 and CIFAR-100, in comparisons with other approaches without data augmentation.

As both stochastic pooling (Zeiler & Fergus, 2013) and max-pooling dropout randomly sample activation based on multinomial distributions at pooling stage, it becomes interesting to compare their performance. Experimental results show that stochastic pooling performs between max-pooling dropout with different retaining probabilities, yet max-pooling dropout with typical retaining probabilities often outperforms stochastic pooling by a large margin.

In this paper, dropout on the input to max-pooling layers is also called max-pooling dropout for brevity. Similarly, dropout on the input to convolutional (or fully-connected) layers is called convolutional (or fully-connected) dropout.

## 2   Review of Dropout Training for Convolutional Neural Networks

CNNs have far been known to produce remarkable performance on MNIST (LeCun et al., 1998), but they, together with other neural network models, fell out of favor in practical machine learning as simpler models such as SVMs became the popular choices in the 1990s and 2000s. With deep learning renaissance (Hinton & Salakhutdinov, 2006; Ciresan, Meier, & Schmidhuber, 2012; Bengio, Courville, & Vincent, 2013), CNNs regained attentions from machine learning and computer vision community. Like other deep models, many issues can arise with deep CNNs if they are naively trained. Two main issues are computation time and over-fitting. Regarding the former problem, GPUs help a lot by speeding up computation significantly.

To combat over-fitting, a wide range of regularization techniques have been developed. A simple but effective method is adding $l_2$ penalty to the network weights. Other common forms of regularization include early stopping, Bayesian fitting (Mackay, 1995), weight elimination (Ledoux & Talagrand, 1991) and data augmentation. In practice, employing these techniques when training big neural networks provides better test performances than smaller networks trained without any regularization.

Dropout is a new regularization technique that has been more recently employed in deep learning. It is similar to bagging (Breiman, 1996), in which a set of models are trained on different subsets of the same training data. At test time, different models' predictions are averaged together. In traditional bagging, each model has independent parameters, and all members would be trained explicitly. In the case of dropout training, there are exponentially many possibly trained models, and these models share the same parameters, but not all of them are explicitly trained. Actually, the number of explicitly trained models is not larger than $m \times e$, where *m* is the number of training example, and *e* is the training epochs. This is much smaller than the number of possibly trained models, $2^n$ ( *n* is number of hidden units in a feed-forward neural networks). Therefore, a vast majority of models are not explicitly trained at training time.

At test time, bagging makes a prediction by averaging together all the sub-models' predictions with the arithmetic mean, but it is not obvious how to do so with the exponentially many models trained by dropout. Fortunately, the average prediction of exponentially many sub-models can be approximately computed simply by running the whole network with the weights scaled by retaining probability. The approximation has been mathematically characterized for linear and sigmoidal networks (Baldi & Sadowski, 2014; Wager el al., 2013); for piecewise linear networks such as rectified linear networks, Warde et al. (2014) empirically showed that weight-scaling approximation is a remarkable and accurate surrogate for the true geometric mean, by comparing against the true average in small enough networks that the exact computation is tractable.

Since dropout was thought to be far less advantageous in convolutional layers, pioneering work by Hinton et al. (2012) only applied it to fully-connected layers. It was the reason they provided that the convolutional shared-filter architecture was a drastic reduction in the number of parameters and thus reduced its possibility to overfit in convolutional layers. Wonderful work by Krizhevsky et al. (2012) trained a very big convolutional neural net, which had 60 million parameters, to classify 1.2 million high-resolution images of ImageNet into the 1000 different categories. Two primary methods were used to reduce over-fitting in their experiments. The first one was data augmentation, an easiest and most commonly used approach to reduce over-fitting for image data. Dropout was exactly the second one. Also, it was only used in fully-connected layers. In the ILSVRC-2012 competition, their deep convolutional neural net yielded top-5 test error rate of 15.3%, far better than the second-best entry, 26.2%, achieved by shallow learning with hand-craft feature engineering. This was considered as a breakthrough in computer vision. From then on, the community believes that deep convolutional nets not only perform best on simple hand-written digits, but also really work on complex natural images.

Compared to original work on dropout, (Srivastava et al., 2014) provided more exhaustive experimental results. In their experiments on CIFAR-10, using dropout in fully-connected layers reduced the test error from 15.60% to 14.32%. Adding dropout to convolutional layers further reduced the error to 12.61%, revealing that applying dropout to convolutional layers aided generalization. Similar performance gains can be observed on CIFAR-100 and SVHN. Still, they did not explore max-pooling dropout.

Stochastic pooling (Zeiler & Fergus, 2013) is a dropout-inspired regularization method. The authors replaced the conventional deterministic pooling operations with a stochastic procedure. Instead of always capturing the strongest activity within each pooling region as max-pooling does, stochastic pooling randomly picks the activations according to a multinomial distribution. At test time, probability weighting is used as an estimate to the average over all possible models. Interestingly, stochastic pooling resembles the case of using dropout in max-pooling layers, so it is worth comparing them.

DropConnect (Wan et al., 2013) is a natural generalization of dropout for regularizing large feed-forward nets. Instead of setting to zero the activations, it sets a randomly picked subset of weights within the network to zero with probability $1 - p$. In other words, the fully-connected layer with DropConnect becomes a sparsely connected layer in which the connections are chosen stochastically during training. Each unit thus only receives input from a random subset of units in the previous layer. DropConnect resembles dropout as it involves stochasticity within the model, but differs in that the stochasticity is on the weights, rather than the output vectors of a layer. Results on several visual recognition datasets showed that DropConnect often outperformed dropout.

Maxout network (Goodfellow et al., 2013) is another model inspired by dropout. The maxout unit picks the maximum value within a group of linear pieces as its activation. This type of nonlinearity is a generalization of rectified activation function and is capable of approximating arbitrary convex function. Combining with dropout, maxout networks have been shown to achieve best results on MNIST, CIFAR-10, CIFAR-100 and SVHN. However, the authors did not train maxout networks without dropout. Besides, they did not train the rectified counterparts with dropout and directly compare it with maxout networks. Therefore, it was not clear that which factor contributed to such remarkable results.

## 3 Max-Pooling Dropout and Convolutional Dropout

We now demonstrate that max-pooling dropout is equivalent to sampling activation according to a multinomial distribution at training time. Basing on this interpretation, we propose to use probabilistic weighted pooling at test time. We also describe convolutional dropout.

### 3.1 Max-Pooling Dropout

Consider a standard CNN composed of alternating convolutional and pooling layers, with fully-connected layers on top. On each presentation of a training example, if layer *l* is followed by a pooling layer, the forward propagation without dropout can be described as

$$a_j^{(l+1)} = pool(a_1^{(l)},..., a_i^{(l)},..., a_n^{(l)}), i \in R_j^{(l)}. \tag{1}$$

Here $R_j^{(l)}$ is pooling region *j* at layer *l* and $a_i^{(l)}$ is the activity of each neuron within it. $n = |R_j^{(l)}|$ is the number of units in $R_j^{(l)}$. *Pool*() denotes the pooling function. Pooling operation provides a form of spatial transformation invariance as well as reduces the computational complexity for upper layers. An ideal pooling method is expected to preserve task-related information while discarding irrelevant image details. Two popular choices are average- and max-pooling. Average-pooling takes all activations in a pooling region into consideration with equal contributions. This may downplay high activations as many low activations are averagely included. Max-pooling only captures the strongest activation, and disregards all other units in the pooling region. We now show that employing dropout in max-pooling layers avoids both disadvantages by introducing stochasticity.

### 3.1.1 Max-Pooling Dropout at Training Time

With dropout, the forward propagation becomes

$$\hat{a}^{(l)} \sim m^{(l)} * a^{(l)}, \tag{2}$$

$$a_j^{(l+1)} = pool(\hat{a}_1^{(l)},..., \hat{a}_i^{(l)},..., \hat{a}_n^{(l)}), i \in R_j^{(l)}. \tag{3}$$

Here $*$ denotes element wise product and $m^{(l)}$ is a binary mask with each element $m_i^{(l)}$ drawn independently from a Bernoulli distribution. This mask is multiplied with activations $a^{(l)}$ in a pooling region at layer *l* to produce dropout-modified activations $\hat{a}^{(l)}$. The modified activations are then passed to pooling layers.

Fig. 1 presents a concrete example to illustrate the effect of dropout in max-pooling layers. Clearly, without dropout, the strongest activation in a pooling regions is always selected as the pooled activation. With dropout, it is not necessary that the strongest activation being the output. Therefore, max-pooling at training time becomes a stochastic procedure.

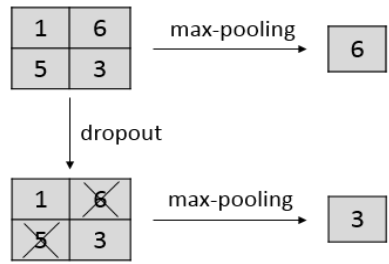

Fig. 1. An illustrating example showing the procedure of max-pooling dropout. The activation in the pooling region is 1, 6, 5 and 3 respectively. Without dropout, the strongest activation 6 is always selected as the output. With dropout, each unit in the pooling region could be possibly dropped out. In this example, only 1 and 3 are retained, then 3 will be the pooled output.

To formulate such stochasticity, suppose the activations $(a_1^{(l)}, a_2^{(l)},..., a_n^{(l)})$ in each pooling region *j* are reordered in non-decreasing order, i.e., $0 \le a_1'^{(l)} \le a_2'^{(l)} ... \le a_n'^{(l)}$.[1] With dropout, each unit in the pooling region could be possibly set to zero with probability of *q* (*q* = 1 – *p* is the dropout probability, and *p* is the retaining probability). As a result, $a_i'^{(l)}$ will be selected as the pooled activation on condition that (1) $a_{i+1}'^{(l)}, a_{i+2}'^{(l)},..., a_n'^{(l)}$ are dropped out, and (2) $a_i'^{(l)}$ is retained. This event occurs with probability of $p_i$ according to *probability theory*:

---

[1] We use rectified linear unit as the activation function, so activations in each pooling region are non-negative. *Sigmoidal* and *tanh* nonlinearities are not adopted due to gradient vanishing effect.

$$\Pr(a_j^{(l+1)} = a_i'^{(l)}) = p_i = pq^{n-i}, (i=1,2,...,n). \tag{4}$$

A special event occurring with probability of $p_0$ ($p_0 = q^n$) is that all the units in a pooling region is dropped out, and the pooled output becomes $a_0'^{(l)}$ ($a_0'^{(l)} = 0$). Therefore, performing max-pooling over the dropout-modified pooling region is exactly sampling from the following multinomial distribution to select an index $i$, then the pooled activation is simply $a_i'^{(l)}$:

$$a_j^{(l+1)} = a_i'^{(l)}, \text{ where } i \sim Multinomial(p_0, p_1, p_2, ..., p_n). \tag{5}$$

Let $s$ be the size of a feature map at layer $l$ (with $r$ feature maps), and $t$ be the size of pooling regions. The number of pooling region is therefore $rs/t$ for non-overlapping pooling. Each pooling region provides $t+1$ choices of the indices, then the number of possibly trained models $C$ at layer $l$ is

$$C = (1+t)^{rs/t} = \left(\sqrt[t]{1+t}\right)^{rs} = b(t)^{rs}. \tag{6}$$

So the number of possibly max-pooling dropout trained models is exponential in the number of units that are fed into max-pooling layers, and the base $b(t)$ ($1 < b(t) = \sqrt[t]{1+t} \leq 2$) depends on the size of pooling regions. Obviously, with the increase of the size of pooling regions, the base $b(t)$ decreases, and the number of possibly trained models becomes smaller. Note that the number of possibly fully-connected dropout trained models is also exponential in the number of units that are fed into fully-connected layers, but with 2 as the base.

### 3.1.2 Probabilistic Weighted Pooling at Test Time

Using dropout in fully-connected layers during training, the whole network containing all the hidden units should be used at test time, but with their outgoing weights halved (Hinton et al. 2012), or with their activations halved. Using max-pooling dropout during training, one might intuitively pick as output the strongest activation scaled down by the retaining probability:

$$a_j^{(l+1)} = p \times \max(a_1^{(l)}, ..., a_i^{(l)}, ..., a_n^{(l)}). \tag{7}$$

We call this pooling scheme *scaled max-pooling*.

Instead we propose to use probabilistic weighted pooling to efficiently get a more accurate approximation of averaging all possibly trained dropout networks. In this pooling scheme, the pooled activity is linear weighted summation over activations in each region:

$$a_j^{(l+1)} = \sum_{i=0}^{n} p_i a_i'^{(l)} = \sum_{i=1}^{n} p_i a_i'^{(l)}. \tag{8}$$

Here $p_i$ is exactly the probability calculated by Eqn. (4). If a unit in a pooling region is selected as output with probability $p_i$ during training, the activation of that unit is weighted by $p_i$ at test time. This ensures that the pooled output at test time is the same with the expected output under the multinomial distribution used to select units at training time.

This type of probabilistic weighted pooling can be interpreted as a form of model averaging where each selection of index $i$ corresponds to a different model. At training time, sampling from multinomial distribution to select a new index produces a new model for each presentation of each training case. The number of possible models is exponential in the number of units that are fed into max-pooling layers. At test time, using probabilistic weighting pooling instead of sampling, we effectively get an approximation of averaging over all of these possible models without instantiating them. Empirical evidence will confirm that probabilistic weighted pooling is a more accurate approximation of averaging all possible dropout models than scaled max-pooling.

### 3.2 Convolutional Dropout

If layer $l$ is followed by a convolutional layer, the forward propagation with dropout is formulated as

$$m_k^{(l)}(i) \sim Bernoulli(p), \tag{9}$$

$$\hat{a}_k^{(l)} = a_k^{(l)} * m_k^{(l)}, \tag{10}$$

$$z_j^{(l+1)} = \sum_{k=1}^{n^{(l)}} \text{conv}(W_j^{(l+1)}, a_k^{(l)}), \tag{11}$$

$$a_j^{(l+1)} = f(z_j^{(l+1)}). \tag{12}$$

Here $a_k^{(l)}$ denotes the activations of feature map $k$ ($k=1,2,…,n^{(l)}$) at layer $l$. The mask matrix $m_k^{(l)}$ consists of independent Bernoulli variables $m_k^{(l)}(i)$. This mask is sampled and multiplied with activations in $k$-th feature map at layer $l$, to produce dropout-modified activations $\hat{a}_k^{(l)}$. These modified activations are convolved with filter $W_j^{(l+1)}$ to produce convolved features $z_j^{(l)}$. The function $f()$ is applied element wise to the convolved features to get the activations of convolutional layers $a_j^{(l+1)}$.

Let $s^2$ be the size of a feature map at layer $l$ (with $r$ feature maps), and $t^2$ be the size of filters, then the number of convolved features is $r \times (s-t+1) \times (s-t+1)$. Each convolved feature is the scalar product of a filter's weights and a local region's activations, so the number of possibly different convolved results is $2^{t \times t}$ when applying dropout to layer $l$. Therefore, the number of total possibly sets of convolved features C decided by layer $l$ is

$$C = 2^{t \times t \times r \times (s-t+1) \times (s-t+1)} = 2^{r \times t^2 \times (s-t+1)^2}. \tag{13}$$

Similarly, dropout is turned off at test time. The whole network containing all the units at layer $l$ should be used, but with the filter's weights scaled by the retaining probability. This efficiently gets an estimate of averaging all possibly trained dropout networks. One may expect that convolutional dropout helps generalization by reducing over-fitting. However, it is far less advantageous, since the shared-filter and local-connectivity architecture in convolutional layers is a drastic reduction in the number of parameters and this already reduces the possibility to overfit (Hinton et al., 2012). Our empirical results will confirm this point that it does improve the generalization to test data, but is often inferior to max-pooling or fully-connected dropout.

## 4 Empirical Evaluations

The purpose of our experiments is threefold: (1) to provide empirical evidence that probabilistic weighted pooling is a more accurate approximation of averaging all possibly max-pooling dropout trained models than max-pooling and scaled max-pooling, (2) to explore dropout training in different layers and (3) to compare max-pooling dropout against stochastic pooling.

We use rectified linear function (Krizhevsky, Sutskever, & Hinton, 2012; Vinod & Hinton, 2010) for convolutional and fully-connected layers, and softmax activation function for the output layer. More commonly used *sigmoidal* and *tanh* nonlinearities are not adopted due to gradient vanishing problem with them. Vanishing gradient effect causes slow optimization convergence, yet training a deep model is already very computationally expensive. Local response normalization is applied after applying rectified non-linearity in certain layers (Krizhevsky, Sutskever, & Hinton, 2012).

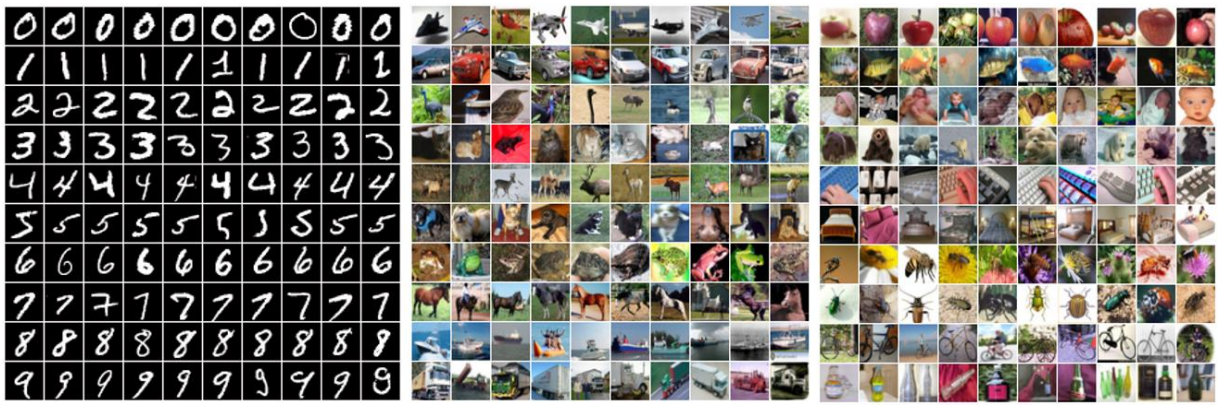

(a) MNIST          (b) CIFAR-10          (c) CIFAR-100

Fig. 2. Example images from MNIST, CIFAR-10 and CIFAR-100. Each row corresponds to a different category. For CIFAR-100, only images of ten categories are displayed.

We train our models using stochastic mini-batch gradient descent with a batch size of 100, momentum of 0.95, learning rate of 0.1 to minimize the cross entropy loss. The weights in all layers are initialized from a zero-mean Gaussian distribution with 0.1 as standard deviation and the constant 0 as the neuron biases in all layers. The heuristic (Hinton et al., 2013) which we follow is to reduce the learning rate twice by a factor of ten before terminating the training. We retain each unit with probability of 0.5 (i.e. $p = 0.5$) by default. Specially, $p$ is set to 0.8 for the first fully-connected layer. Experiments are conducted on three widely used datasets: MNIST, CIFAR-10 and CIFAR-100. Fig. 2 displays the example images.

The CNN architecture in our experiments is denoted in the following way: 2x32x32-32C5-3P2-64C5-3P2-1000N-10N represents a CNN with 2 input image of size 32x32, a convolutional layer with 32 feature maps and 5x5 filters, a pooling layer with pooling region 3x3 and stride 2, a convolutional layer with 64 feature maps and 5x5 filters, a pooling layer with pooling region 3x3 and stride 2, a fully-connected layer with 1000 hidden units, and an output layer with 10 units (one per class).

## 4.1  MNIST

We initially conduct experiments using MNIST, a widely used benchmark dataset in computer vision. It consists of 28x28 pixel grayscale images, each containing a digit 0 to 9. There are 60,000 training and 10,000 test examples. We do not perform any pre-processing except scaling the pixel values to [0, 1].

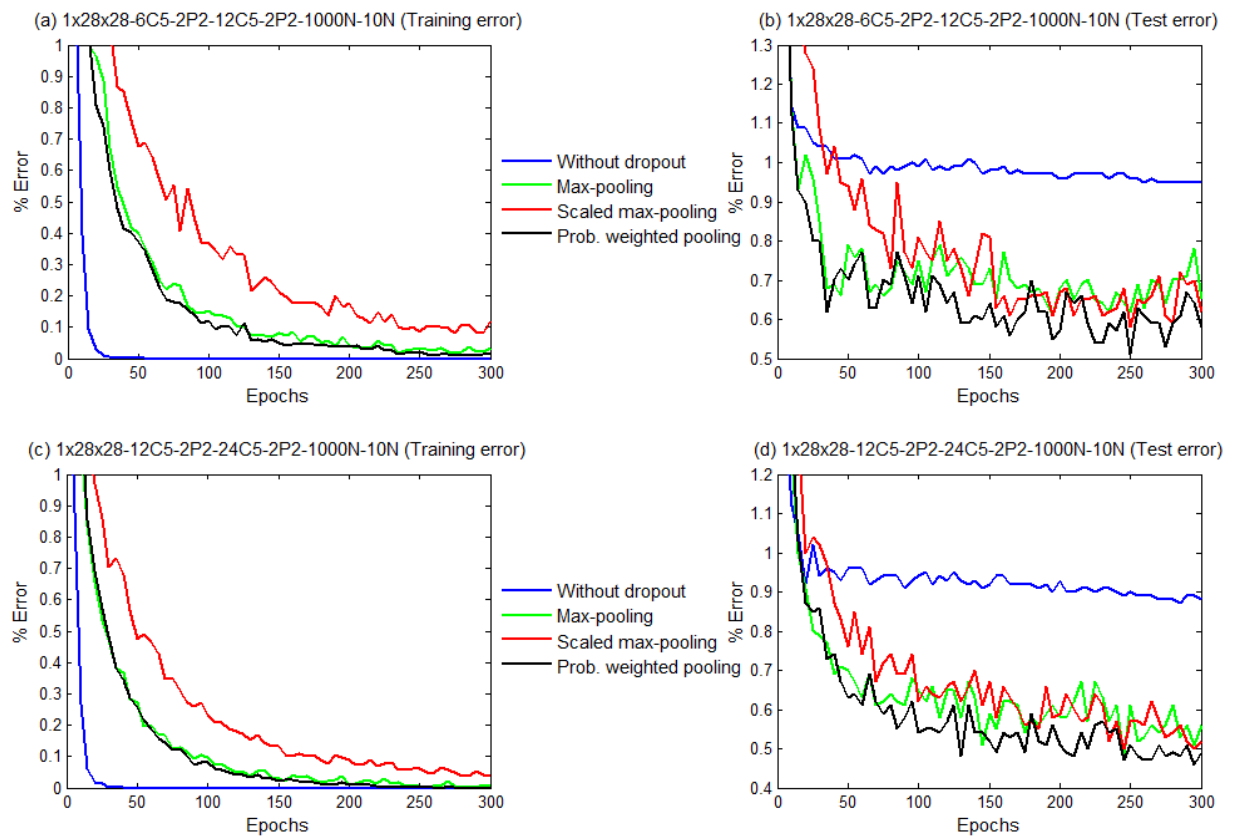

Fig. 3. MNIST training and test errors for different pooling methods at test time. Max-pooling dropout is used during training. Max-pooling without dropout is presented as the baseline. (a) and (b) illustrate the training and test errors produced by the smaller architecture, 1x28x28-6C5-2P2-12C5-2P2-1000N-10N. (c) and (d) illustrate the training and test errors produced by the bigger one, 1x28x28-12C5-2P2-24C5-2P2-1000N-10N.

### 4.1.1 Probabilistic Weighted Pooling vs. Scaled Max-Pooling

To validate the superiority of probabilistic weighted pooling over max-pooling and scaled max-pooling, we first train different CNN models using two different architectures, 1x28x28-6C5-2P2-12C5-2P2-1000N-10N and 1x28x28-12C5-2P2-24C5-2P2-1000N-10N. Max-pooling dropout is applied during training. At test time, max-pooling, scaled max-pooling and probabilistic weighted pooling is used respectively to act as model averaging.

Fig. 3 illustrates the training and test error rates of both CNN architectures over 300 training epochs. Using probabilistic weighted pooling at test time, the training errors are always lower than those produced by scaled max-pooling. This is a good news that probabilistic weighted pooling fits the training data better with the same model parameters. In other words, probabilistic weighted pooling is a more accurate estimation of averaging all possibly trained dropout models than scaled max-pooling. On test data, both probabilistic weighted pooling and scaled max-pooling produces lower errors compared to max-pooling without dropout. Probabilistic weighted pooling generally produces lower errors than scaled max-pooling. This indicates that probabilistic weighted pooling not only fits the training data better, but also generalizes better to the test data. In terms of training performance, max-pooling is about the same with probabilistic weighted pooling, but probabilistic weighted pooling clearly outperforms max-pooling on the test data.

We then train different CNN models with different retaining probabilities for max-pooling dropout. The CNN architecture 1x28x28-20C5-2P2-40C5-2P2-1000N-10N is trained for 1000 epochs. Fig. 4 compares the test performance produced by different pooling methods at test time. Generally, probabilistic weighted pooling performs better than max-pooling and scaled max-pooling with different retaining probabilities. For small $p$, max-pooling and scaled max-pooling performs very poorly; probabilistic weighted pooling is considerably better. With the increase of $p$, the performance gap becomes smaller. This is not surprising as the pooled outputs for different pooling methods are close to each other for large $p$. An extreme case is that when $p = 1$, scaled max-pooling and probabilistic weighted pooling are exactly the same with max-pooling.

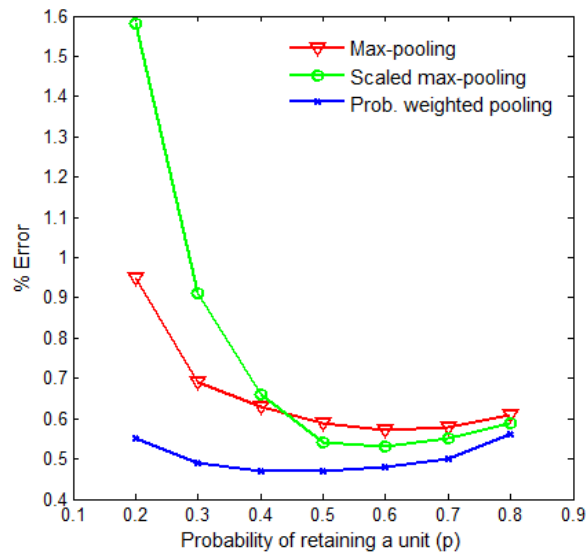

Fig. 4. MNIST test errors for different pooling methods at test time. Max-pooling dropout is used to train CNN models with different retaining probabilities at training time.

### 4.1.2 Dropout Training in Different Layers

To investigate the effect of using dropout in different layers, we first train different models by separately using dropout on the input to convolutional, max-pooling and fully-connected layers. Models are trained using two CNN architectures: 1x28x28-6C5-2P2-12C5-2P2-1000N-10N and 1x28x28-12C5-2P2-24C5-2P2-1000N-10N. The case without dropout is used as the baseline. For max-pooling dropout, only probabilistic weighted pooling is used at test time for its superiority over max-pooling and scaled max-pooling.

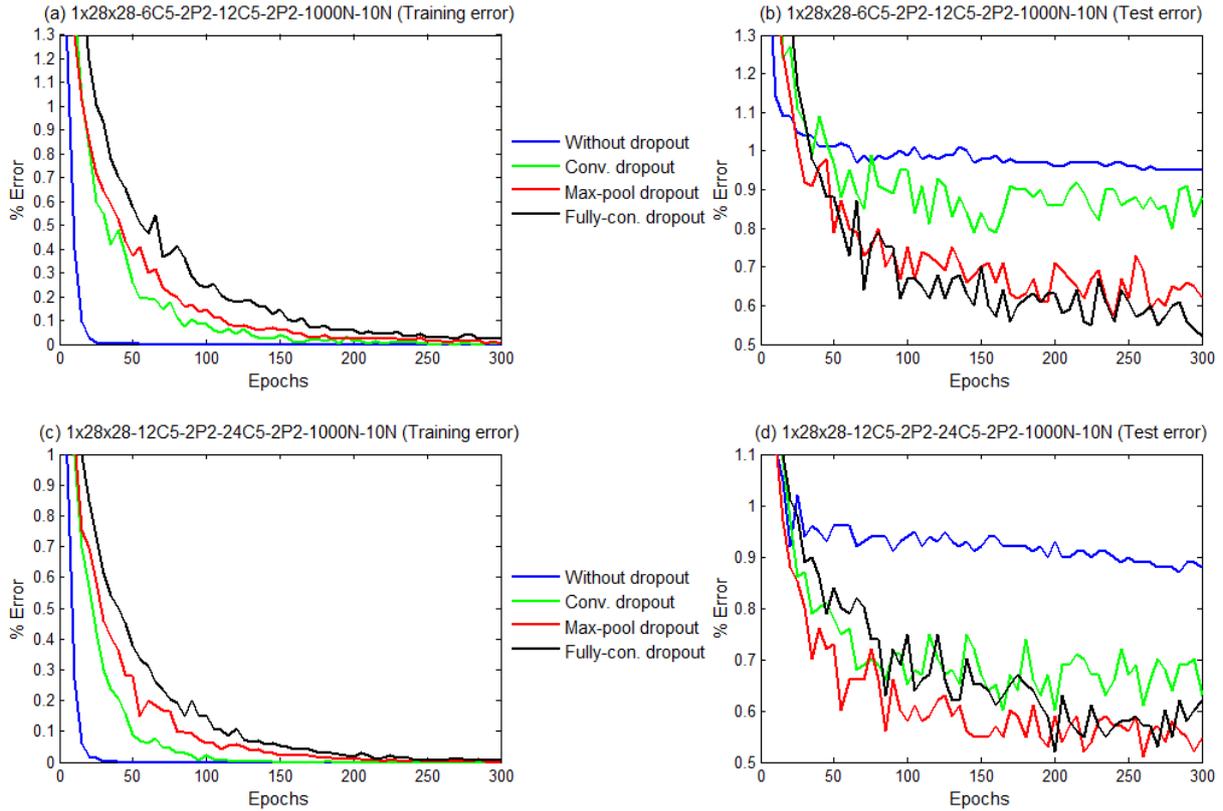

Fig. 5. MNIST training and test errors for separately using convolutional, max-pooling and fully-connected dropout. The case without dropout is also presented for comparison's purpose. (a) and (b) illustrate the training and test errors produced by the smaller architecture, 1x28x28-6C5-2P2-12C5-2P2-1000N-10N. (c) and (d) illustrate the training and test errors produced by the bigger one, 1x28x28-12C5-2P2-24C5-2P2-1000N-10N.

Fig. 5 illustrates the progression of these models' training and test error rates over 300 epochs. Without dropout, the training error falls rapidly to zero, but the test error stands at about 0.9% for both CNN architectures, revealing that both models suffer from severe over-fitting. With convolutional dropout, the training error decreases slower, and the test error drops to 0.85% and 0.65% respectively for the above two CNN architectures. This indicates that convolutional dropout can reduce over-fitting and aid generalization to the test data. With max-pooling dropout, the training error also falls down slower while the test error decreases to about 0.60% and 0.53% respectively, much better than convolutional dropout. With fully-connected dropout, the training error still decreases much slower, and the performance on test data is much better compared to the case without dropout.

Note that fully-connected dropout is the best performer on the smaller architecture. This is not surprising since fully-connected layers have many units and need regularization; the convolutional and pooling component is thin enough, so regularization is relatively not so advantageous. Max-pooling dropout performs well on both architectures, even though the CNN architectures are small, indicating that max-pooling dropout is a robust regularizer. Convolutional dropout is not trivial compared to the case without dropout, but its performance gain is relatively small compared to max-pooling dropout and fully-connected dropout. This is to be expected, since the shared-filter and local-connectivity convolutional architecture is a drastic reduction in the number of parameters and this already reduces the possibility to overfit.

| Method | Error % |
|---|---|
| No dropout | 0.81 |
| Fully-connected dropout | 0.56 |
| Convolutional dropout | 0.60 |
| Max-pooling dropout | 0.47 |
| Convolutional and fully-connected dropout | 0.50 |
| Convolutional and max-pooling dropout | 0.61 |
| Max-pooling and fully-connected dropout | **0.39** |
| NIN + dropout (Lin, Chen, & Yan, 2014) | 0.47 |
| Maxout + dropout (Goodfellow et al., 2013) | **0.45** |
| Stochastic pooling (Zeiler & Fergus, 2013) | 0.47 |

Table 1. MNIST test errors for 1x28x28-20C5-2P2-40C5-2P2-1000N-10N trained with dropout in various types of layers compared to current state-of-the-art CNN models, excluding methods that augment the training data.

As observed, separately using convolutional, max-pooling and fully-connected dropout reduces over-fitting and improves test performance. To achieve better results, we try using dropout simultaneously on the input to different layers. For max-pooling dropout, only probabilistic weighted pooling is used at test time.

To this end, we train various models for 1000 training epochs using architecture 1x28x28-20C5-2P2-40C5-2P2-1000N-10N. For comparison's purpose, the results of separately using dropout in different layers are also presented. Table 1 records these models' test errors, as well as recently published state-of-the-art results produced by CNNs. The best performing CNN model that do not use dropout yields an error rate of 0.81% on MNIST. With fully-connected dropout, the test error decreases to 0.56%. Convolutional dropout reduces the test error to 0.60%. With max-pooling dropout, the test error drops to 0.47%. However, Compared to only using convolutional or max-pooling dropout, simultaneously using convolutional and pooling dropout does not decrease, but increase the error. Simultaneously employing max-pooling and fully-connected dropout, we obtain test error rate of 0.39%. This beats the previous state-of-the-art results to the best of our knowledge, 0.47% (Lin, Chen, & Yan, 2014), 0.45% (Goodfellow et al., 2013) and 0.47% (Zeiler & Fergus, 2013).

### 4.2 CIFAR-10

The CIFAR-10 dataset (Krizhevsky, 2009) consists of ten classes of natural images with 50,000 examples for training and 10,000 for testing. Each example is a 32x32 RGB image taken from the tiny images dataset collected from the web. We also scale to [0, 1] for this dataset and subtract the mean value of each channel computed over the dataset for each image.

Compared to MNIST, the CIFAR-10 images are highly varied within each class, so we train deeper and wider nets to model the complex non-linear relationships. The architecture is 3x32x32-96C5-3P2-128C3-3P2-256C3-3P2-2000N-2000N-10N. A notable difference from the architecture used for MNIST is that each input image for CIFAR-10 has three channels, with each channel corresponding to an R/G/B component. Models are trained for 1000 epochs.

We first compares different pooling methods at test time for max-pooling dropout trained models on CIFAR-10. The retaining probability is set to 0.3, 0.5 and 0.7 respectively. At test time, max-pooling, scaled max-pooling and probabilistic weighted pooling are respectively used to act as model averaging. Fig. 6 presents the test performance of these pooling methods. Again, for small retaining probability $p = 0.3$, scaled max-pooling and probabilistic weighted pooling perform poorly. Probabilistic weighted pooling is the best performer with different retaining probabilities. The increase of $p$ narrows different pooling methods' performance gap.

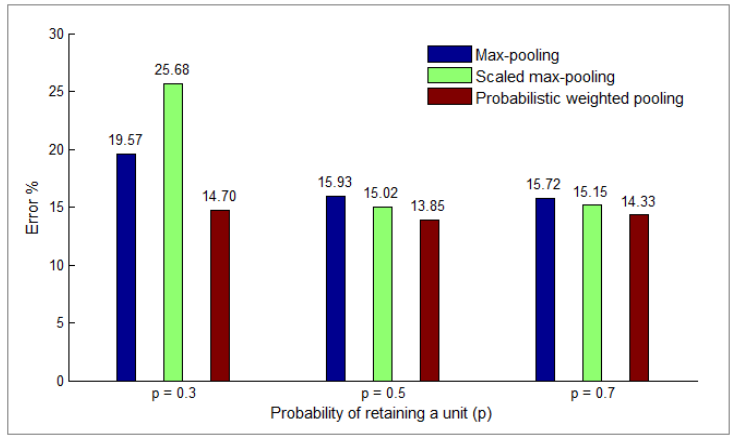

Fig. 6. CIFAR-10 test errors for different pooling methods at test time. Max-pooling dropout is used to train CNN models with different retaining probabilities at training time.

We also train various CNN models by separately and simultaneously using dropout in different layers on CIFAR-10. For max-pooling dropout trained models, only probabilistic weighted pooling is used at test time. Table 2 records the performance of various models, as well as recently published results produced by CNN models. Without dropout, the best test error is 16.50%. Separately using convolutional, max-pooling and fully-connected dropout reduces the test error. Convolutional dropout still underperforms max-pooling dropout. Again, simultaneously using convolutional and max-pooling dropout seems to easily result in too strong regularization, and thus decreases the test performance. Using max-pooling and fully-connected dropout, we obtain very competitive test error rate, 11.29%, compared to recently published results, 11.35% (Springenberg & Riedmiller, 2014), 10.41% (Lin, Chen, & Yan, 2014), 11.68% (Goodfellow et al., 2013), 15.14% (Zeiler & Fergus, 2013) and 15.60% (Hinton et al., 2012).

| Method | Error % |
|---|---|
| No dropout | 16.50 |
| Fully-connected dropout | 14.24 |
| Convolutional dropout | 15.78 |
| Max-pooling dropout | 13.85 |
| Convolutional and max-pooling dropout | 15.87 |
| Convolutional and fully-connected dropout | 12.93 |
| Max-pooling and fully-connected dropout | 11.29 |
| Probout (Springenberg & Riedmiller, 2014) | 11.35 |
| NIN + dropout (Lin, Chen, & Yan, 2014) | 10.41 |
| Maxout + dropout (Goodfellow et al., 2013) | 11.68 |
| Stochastic pooling (Zeiler & Fergus, 2013) | 15.14 |
| Dropout the last hidden layer (Hinton et al., 2012) | 15.60 |

Table 2. CIFAR-10 test errors for dropout training in various types of layers compared to current state-of-the-art CNN models, excluding methods that augment the training data.

### 4.3 CIFAR-100

CIFAR-100 (Krizhevsky, 2009) is just like CIFAR-10, but with 100 categories. We also scale to [0, 1] for CIFAR-100 and subtract the mean value of each R/G/B channel. The architecture is the same with that of CIFAR-10 except that the number of output units is 100. Models are trained for 1000 epochs.

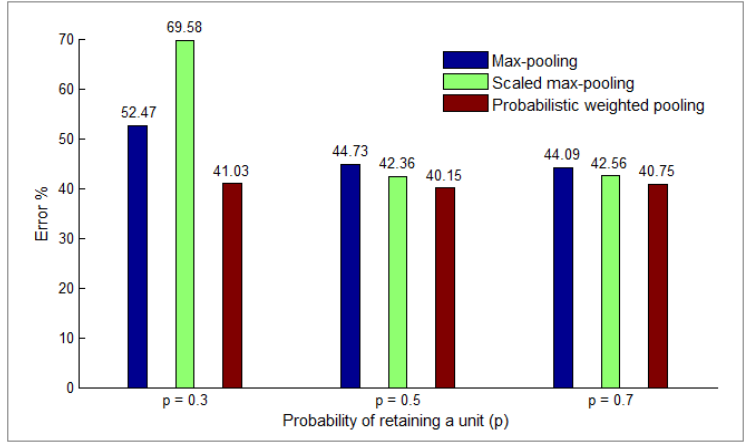

Fig. 7. CIFAR-100 test errors for different pooling methods at test time. Max-pooling dropout is used to train CNN models with different retaining probabilities at training time.

Fig. 7 compares different pooling methods at test time for max-pooling dropout trained models on CIFAR-100. The retaining probability is set to 0.3, 0.5 and 0.7 respectively. Again, probabilistic weighted pooling shows clear superiority over max-pooling and scaled max-pooling.

We then train various CNN models by separately and simultaneously using dropout in different layers on CIFAR-100. For max-pooling dropout trained models, only probabilistic weighted pooling is used at test time. As shown in Table 3, it improves the generalization to test data that using dropout separately in convolutional, max-pooling and fully-connected layers. Again, if dropout is employed improperly, the performance would decrease, such as simultaneously using convolutional and max-pooling dropout. The best result, 37.13%, is achieved using max-pooling and fully-connected dropout.

| Method | Error % |
| --- | --- |
| No dropout | 44.47 |
| Fully-connected dropout | 41.18 |
| Convolutional dropout | 42.23 |
| Max-pooling dropout | 40.15 |
| Convolutional and max-pooling dropout | 43.33 |
| Convolutional and fully-connected dropout | 38.56 |
| Max-pooling and fully-connected dropout | **37.13** |
| Probout ( Springenberg & Riedmiller, 2014) | 38.14 |
| NIN + dropout (Lin, Chen, & Yan, 2014) | **35.68** |
| Maxout + dropout (Goodfellow et al., 2013) | 38.57 |
| Stochastic pooling (Zeiler & Fergus, 2013) | 42.51 |

Table 3. CIFAR-100 test errors for dropout training in various types of layers compared to current state-of-the-art CNN models, excluding methods that augment the training data.

### 4.4 Max-Pooling Dropout and Stochastic Pooling

Similar to max-pooling dropout, stochastic pooling (Zeiler & Fergus, 2013) also randomly picks activation according to a multinomial distribution at training time, and also involves probabilistic weighting at test time. More concretely, at training time it first computes the probability $p_i$ for each unit within pooling region $j$ at layer $l$ by normalizing the activations:

$$p_i = \frac{a_i^{(l)}}{\sum_{k=1}^{n} a_k^{(l)}}, (i = 1, 2, ..., n). \tag{14}$$

It then samples from a multinomial distribution based on $p_i$ to select an index $i$ in the pooling region. The pooled activation is simply $a_i^{(l)}$:

$$a_j^{(l+1)} = a_i^{(l)}, \text{ where } i \sim Multinomial(p_1, p_2, ..., p_n). \tag{15}$$

At test time, probabilistic weighting is adopted to act as model averaging. That is, the activations in each pooling region are weighted by the probability $p_i$ and summed:

$$a_j^{(l+1)} = \sum_{i=1}^{n} p_i a_i^{(l)}. \tag{16}$$

One may have found that stochastic pooling bears much resemblance to max-pooling dropout, as both involve stochasticity at pooling stage. We are therefore very interested in their performance differences. As presented in Sec. 3, for max-pooling dropout, the number of possibly trained models at layer $l$ is $b(t)^{rs}$, where $b(t) = \sqrt[t]{(t+1)}$. For stochastic pooling, since each pooling region provides $t$ choices, the number of possibly trained models $C$ at layer $l$ is

$$C = t^{rs/t} = \left(\sqrt[t]{t}\right)^{rs} = b(t)^{rs}. \tag{17}$$

So the number of possibly trained models for stochastic pooling is also exponential in the number of hidden units, $rs$, but with a different base $b(t)$ $(1 \leq b(t) = \sqrt[t]{t} \leq \sqrt[3]{3})$. Fig. 8 compares this function against the base of max-pooling dropout. Now consider a concrete example. For a certain layer $l$, let $s = 32 \times 32$, $r = 96$, and $t = 2 \times 2$. For stochastic pooling, $b(t) = (2 \times 2)^{1/(2 \times 2)} = 4^{1/4}$; for max-pooling dropout, $b(t) = (1 + 2 \times 2)^{1/(2 \times 2)} = 5^{1/4}$. Max-pooling dropout provides $\varepsilon$ times larger number of possible models, and $\varepsilon$ is computed as follows:

$$\varepsilon = (4/5)^{1/4 \times 32 \times 32 \times 96} = 1.25^{24576}. \tag{18}$$

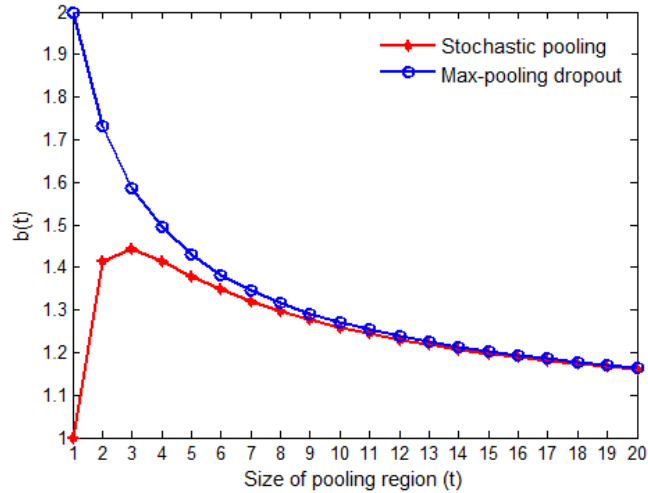

Fig. 8. The base $b(t)$ for stochastic pooling and max-pooling dropout. For max-pooling dropout, $b(t) = \sqrt[t]{1+t}$, while $b(t) = \sqrt[t]{t}$ for stochastic pooling. Here $t$ is the size of pooling regions. The difference in $b(t)$ decreases with the increase of $t$.

Clearly, max-pooling dropout provides extremely larger number of possibly trained models than stochastic pooling, despite the small difference in the base.

To compare their performances, we train various models on MNIST using two CNN architectures, 1x28x28-6C5-2P2-12C5-2P2-1000N-10N and 1x28x28-12C5-2P2-24C5-2P2-1000N-10N. For max-pooling dropout trained models, only probabilistic weighted pooling is used at test time. Fig. 9 illustrates the progression of training and test errors using stochastic pooling and max-pooling dropout with different retaining probabilities over 300 training epochs. Regarding training errors, stochastic pooling falls in between max-pooling dropout with $p = 0.7$ and $p = 0.5$ for both CNN architectures. On the test data, stochastic pooling performs worst for the smaller CNN architecture. For the bigger one, the test performance of stochastic pooling is between max-pooling dropout with $p = 0.7$ and $p = 0.5$.

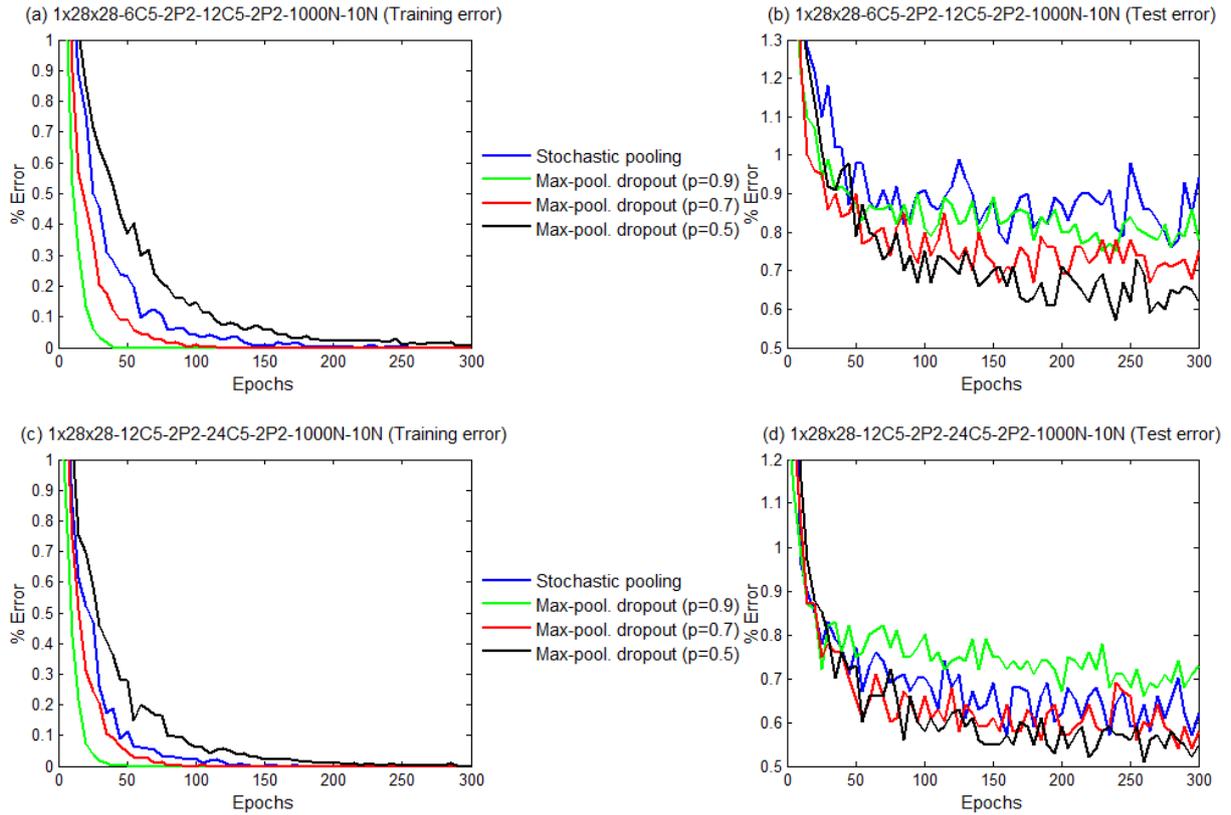

Fig. 9. MNIST training and test errors for stochastic pooling, and max-pooling dropout with different retaining probabilities. (a) and (b) illustrate the training and test errors produced by 1x28x28-6C5-2P2-12C5-2P2-1000N-10N. (c) and (d) illustrate the training and test errors produced by 1x28x28-12C5-2P2-24C5-2P2-1000N-10N.

For a more clear comparison, we train CNN models with more different retaining probabilities on MNIST, CIFAR-10 and CIFAR-100. For max-pooling dropout trained models, only probabilistic weighted pooling is used at test time. Fig. 10 compares the test performances of max-pooling dropout with different retaining probabilities against stochastic pooling. As we can observed, the relation between the performance of max-pooling dropout and the retaining probability *p* is a U-shape. If *p* is too small or too large, max-pooling dropout performs poorer than stochastic pooling. Yet max-pooling dropout with typical *p* (around 0.5) outperforms stochastic pooling by a large margin. Therefore, although stochastic pooling is hyper-parameter free and this saves the tuning of retaining probability, its performance is often inferior to max-pooling dropout.

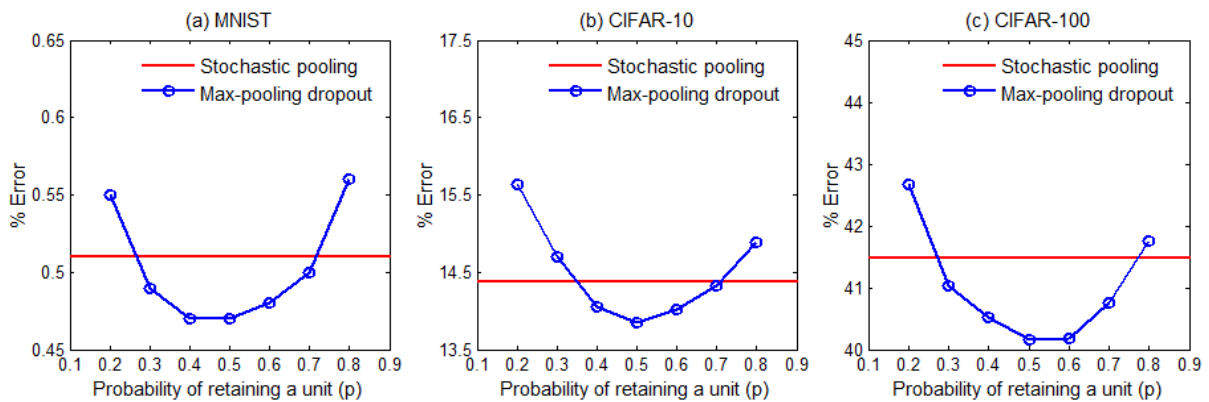

Fig. 10. Test errors for max-pooling dropout with different retaining probabilities against stochastic pooling. The architecture is 1x28x28-20C5-2P2-40C5-2P2-1000N-10N for MNIST, 3x32x32-96C5-3P2-128C3-3P2-256C3-3P2-2000N-2000N-10N for CIFAR-10 and 3x32x32-96C5-3P2-128C3-3P2-256C3-3P2-2000N-2000N-100N for CIFAR-100.

### 4.5  A Brief Summary of Experimental Results

The summary of experimental results is threefold in response to the purpose of our experiments.

(1) For max-pooling dropout trained CNN models, probabilistic weighted pooling not only fits the training data better, but also generalizes better to the test data than max-pooling and scaled max-pooling. For small retaining probability, max-pooling and scaled max-pooling produce very poor results, while probabilistic weighted pooling performs very well. With the increase of the retaining probability, the performance gap becomes smaller.

(2) Separately using convolutional, max-pooling and fully-connected dropout improves performance, but convolutional dropout seems far less advantageous. Simultaneously using dropout in different layers could further reduce the test error, but should be careful. For example, with typical retaining probability 0.5, using convolutional and max-pooling dropout simultaneously does not performs well; while using max-pooling dropout and fully-connected dropout simultaneously achieves best results on MNIST and CIFAR datasets.

It is worth to point out that determining which layers to train with dropout is an open question, since many factors (such as network architecture, the retaining probability and even the training data) have large influence on the performance. Especially, for different network architecture, the best dropout regularization strategy could be quite different. While the best regularization strategy (simultaneously using max-pooling and fully-connected dropout) is consistent across three datasets in our experiments, it is possible that these networks just need lots of regularization in max-pooling and fully-connected layers.

(3) The retaining probability affects max-pooling's performance, and the relation exhibits U-shape. Max-pooling dropout with typical retaining probabilities (around 0.5) often outperforms stochastic pooling by a large margin.

### 5  Conclusions

This paper mainly addresses the problem of understanding and using dropout on the input to max-pooling layers of convolutional neural nets. At training time, max-pooling dropout is equivalent to randomly picking activation according to a multinomial distribution, and the number of possibly trained networks is exponential in the number of input units to the pooling layers. At test time, a new pooling method, probabilistic weighted pooling, is proposed to act as model averaging. Experimental evidence confirms the benefits of using max-pooling dropout, and validates the superiority of probabilistic weighted pooling over max-pooling and scaled max-pooling. Using dropout simultaneously in different layers could further improve the performance, but should be careful. Using max-pooling dropout and fully-connected dropout, this paper reports better results on MNIST, and comparable performance on CIFAR-10 and CIFAR-100, in comparisons with state-of-the-art results that do not use data augmentation. In addition, considering that stochastic pooling is similar to max-pooling dropout, we empirically compare them and show that the performance of stochastic pooling is between those produced by max-pooling dropout with different retaining probabilities.

### Acknowledgements

This work was supported in part by National Natural Science Foundation of China under grant 61371148.

# References

Baldi, P., & Sadowski, P. (2014). The dropout learning algorithm. *Artificial Intelligence*, 210, 78-122.

Bengio, Y., Courville, A., & Vincent, P. (2013). Representation learning: a review and new perspectives. *IEEE Transactions on Pattern Analysis and Machine Intelligence*, 35, 1798-1828.

Boureau, Y. L., Ponce J., & LeCun, Y. (2010). A theoretical analysis of feature pooling in visual recognition. In *Proceedings 27th of International Conference on Machine Learning (ICML 2010)*.

Breiman, L. (1996). Bagging predictors. *Machine Learning*, 24, 123-140.

Ciresan. D., Meier, U., & Schmidhuber, J. (2012). Multi-column deep neural networks for image classification. In *Proceedings of 2014 IEEE Conference on Computer Vision and Pattern Recognition (CVPR 2012)*.

Goodfellow, I. J., Warde-Farley, D., Mirza, M., Courville, A., & Bengio, Y. (2013). Maxout networks. In *Proceedings of 30th International Conference on Machine Learning (ICML 2013)*.

Hinton, G. E., & Salakhutdinov, R. R. (2006). Reducing the dimensionality of data with neural networks. *Science*, 313, 504-507.

Hinton, G. E., Srivastave, N., Krizhevsky, A., Sutskever, I. & Salakhutdinov, R. R. (2012). Improving neural networks by preventing co-adaption of feature detectors. *arXiv 1207.0580*.

Springenberg J. T., & Riedmiller M. (2014). Improving deep neural networks with probabilistic maxout units. In *Proceedings of 3rd International Conference on Learning Representations (ICLR 2014)*.

Krizhevsky, A. (2009). Learning multiple layers of features from tiny images. *M.S. diss., Department of Computer Science, University of Toronto*.

Krizhevsky, A., Sutskever, I., & Hinton, G. E. (2012). ImageNet classification with deep convolutional neural networks. In *Advances in Neural Information Processing Systems (NIPS 2012)*.

LeCun, Y., Bottou, L., Bengio, Y. & Haffner, P. (1998). Gradient-based learning applied to document recognition. In *Proceedings of the IEEE*.

Ledoux, M., & Talagrand, M. (1991). Probability in banach spaces. *Springer*.

Lin, M., Chen, Q., & Yan S. (2014). Network in network. In *Proceedings of 3rd International Conference on Learning Representations (ICLR 2014)*.

Mackay, D. C. (1995). Probable networks and plausible predictions: A review of practical Bayesian methods for supervised neural networks. In *Bayesian Methods for Backpropagation Networks*.

Scherer, D., Muller, A., & Behnke, S. (2010). Evaluation of pooling operations in convolutional architectures for object recognition. In *Proceedings of 20th International Conference on Artificial Neural Networks (ICANN 2010)*.

Srivastava, N., Hinton. G. E., Krizhevsky, A., Sutskever, I., & Salakhutdinov, R. (2014). Dropout: a simple way to prevent neural networks from overfitting. *Journal of Machine Learning Research*, 15, 1929-1958.

Vinod, N., & Hinton, G. E. (2010). Rectified linear units improve restricted Boltzmann machines. In *Proceedings 27th of International Conference on Machine Learning (ICML 2010)*.

Wan, L., Zeiler, M. D., Zhang, S., LeCun, Y., & Fergus, R. (2013). Regularization of neural networks using DropConnect. In *Proceedings of 30th International Conference on Machine Learning (ICML 2013)*.

Warde, F. D., Goodfellow, I. J., Courville, A., & Bengio, Y. (2014). An empirical analysis of dropout in piecewise linear networks. *In Proceedings of 3rd International Conference on Learning Representations (ICLR 2014)*.

Wager, S., Wang, S., & Liang, P. (2013). Dropout training as adaptive regularization. In *Advances in Neural Information Processing Systems (NIPS 2013)*.

Zeiler, M. D., & Fergus R. (2013). Stochastic pooling for regularization of deep convolutional neural networks. In *Proceedings of 2nd International Conference on Learning Representations (ICLR 2013)*.